# Optical Neural Networks

**Grant Fennessy**
Vanderbilt University
Nashville, TN
`grant.fennessy@vanderbilt.edu`

**Yevgeniy Vorobeychik**
Vanderbilt University
Nashville, TN
`yevgeniy.vorobeychik@vanderbilt.edu`

## Abstract

We develop a novel optical neural network (ONN) framework which introduces a degree of scalar invariance to image classification estimation. Taking a hint from the human eye, which has higher resolution near the center of the retina, images are broken out into multiple levels of varying zoom based on a focal point. Each level is passed through an identical convolutional neural network (CNN) in a Siamese fashion, and the results are recombined to produce a high accuracy estimate of the object class. ONNs act as a wrapper around existing CNNs, and can thus be applied to many existing algorithms to produce notable accuracy improvements without having to change the underlying architecture.

## 1 Introduction

A human's ability to rapidly identify an object is second nature, but the task is exceedingly difficult for machines. Traditional algorithms attempted to solve this problem with extremely complicated mathematical regimes, but everything changed with the introduction of AlexNet [5], which utilized convolutional neural networks (CNNs) for object identification, an architecture with a translation invariant approach to feature identification.

Since AlexNet, even deeper models like GoogLe Net [8] have been used in an attempt to improve accuracy and reduce overfitting. An approach that creates preprocessing modules that extract features and transform them before they are processed by the CNN [3] has seen headway recently. Even attention modules for evaluating which features are most "eye-grabbing" [10] are proving successful. All of these models greatly augment the accuracy of raw CNNs, but none aggressively approach the issue of scalar invariance.

Optical neural networks (ONNs) seek to introduce a degree of scalar invariance into the model by loosely emulating how the human eye's resolution is configured: the retina provides high resolution at the center, but that resolution fades off quickly along the radius. In this way, the eye can identify high detail components of an object, while still being able to capture enough peripheral detail to make a single estimate of the object at hand. When observing an elephant, only a small portion of the creature is seen in high detail, while the rest is observed in much lower resolution in the periphery, yet we are still able to easily identify that it is in fact an elephant - we don't have to scan over every detail and subfeature, but can focus on only one point and rapidly make a determination.

The optical neural network architecture emulates this varying resolution to produce relatively fast estimates compared to the size of the input image. Instead of passing the entire image into a bloated CNN, the image is broken out into multiple 'zoom' levels around a focal point on the image. A region from each zoom level is then extracted and passed into a smaller CNN. Using a Siamese arrangement, each image goes through the same processing, and the results are unified into a single estimate. The net result is a higher degree of accuracy over the baseline for the given CNN size. Accuracy can be further improved by merging multiple focal points of one image together to create a single merged estimate.

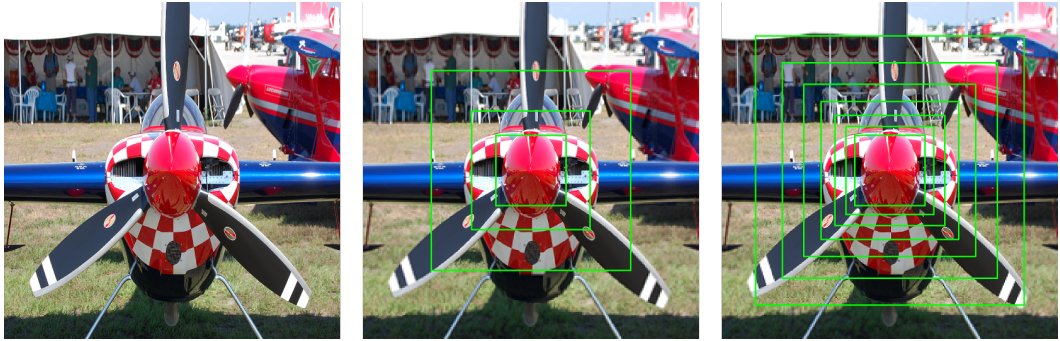

Figure 1: An image of an aircraft, courtesy of OpenImages [4]. The left image is the base 1068x1068 image. The center has extracted 224x224 regions from 4 zoom levels. To visualize how the ONN effect works, the images are scaled up to their effective sizes (as depicted by the green outlines) and overlaid on top of each other. Note that the outer regions are of a lower resolution than the central regions. The right image utilizes 8 zoom levels. Both images make use of a focal point on the center of the image, though the focal point could be anywhere on the image.

The ONN architecture is designed to be modular and expandable, allowing implementers to specify their components to meet their needs, but also providing ample research opportunity for continued improvement. Importantly, the architecture can be overlaid on top of existing CNNs to yield notable accuracy improvements.

We discuss the important research leading up to this paper before explaining the architecture in full detail. The architecture is fairly open-ended, so the specifics of our implementation are provided. We then cover all of the experiments run against the architecture and comment on the results before discussing future work unlocked by the ONN architecture.

## 2 Related Work

The rise of the modern CNN started with the AlexNet [5] model, featuring a 7 layer convolutional neural network design to classify 224x224 images from the ILSVRC dataset [6]. GoogLe Net [8] built an even deeper CNN model based off of AlexNet, featuring 22 layers. ResNet added [1] residual learning into the fray, a technique aimed at helping reduce the problem of vanishing and exploding gradients. Our model is built off of the VGG [7] model, but could actually be applied to any of these models, along with other more advanced CNN approaches.

Jaderberg et al. [3] introduced the Spatial Transformer Networks model, which invokes a touch of spatial invariance. A preprocessor filters images entering the CNN, clipping, transforming, and rotating them to produce an image that is more readily ingestible by the CNN. A variation of the model uses multiple transformer networks in parallel, each trained against specific features from the Caltech Birds dataset [9]. These transformers learn major features of the birds and help convert them into more CNN-friendly images. Nevertheless, the resulting model does not exhibit *scalar* invariance of the kind we are aiming for.

Residual attention modules [10] attempt to find features in images that humans are more likely to look at, allowing the underlying CNN to more rapidly identify features of interest instead of wasting time on background noise (such as trees and skylines). Our focal points are inspired by this idea, but we go significantly beyond in considering specifically the issue of scale invariance.

Dilated residual networks [11] present a solution to a common problem amongst CNNs: low degrees of spatial accuracy, a byproduct of the multiple layers of convolution and pooling that reduce the image size and make feature localization more difficult. This problem could actually be avoided to a certain degree with ONNs, as the accuracy can be retained by using higher zoom levels, allowing a greater degree of fidelity when regressing to a bounding box centroid or feature point.

Squeeze-and-Excitation networks [2] attempt to leverage global information to suppress less useful features, instead focusing on features that provide maximal information about which object is in question. This technique has the potential to be very useful, and can be employed alongside an ONN architecture.

## 3 Architecture

The ONN architecture was designed around the idea of leveraging a focal point and a specified number of zoom levels. The focal point determines where on the image the ONN's "eye" will focus, while the number of zoom levels determines the resolution dropoff rate. Increasing the number of zoom levels will produce a more continuous resolution dropoff.

Though we will define a specific ONN implementation, the architecture is purposefully left open, as there are many ways in which the core principles can be accomplished. Some potential implementations are discussed later in this paper.

Our ONN implementation is a proof of concept of the architecture, leveraging the following principles:

- Take any existing CNN in any existing architecture/model,
- Train the CNN against the focal points and zoom levels, where each extracted image is considered independent. This acts as data augmentation,
- Train a new *unification* network to ingest all zoom levels of a focal point in a Siamese fashion and output a single estimate,
- Combine multiple focal points into a single prediction.

The ONN architecture requires determining a few hyperparameters:

- $F$ is the set of allowed focal points, where a given focal point is declared as $\gamma$, such that $\gamma_x \in [0, 1]$ and $\gamma_y \in [0, 1]$. $F$ can be a subset of all unique focal points, $F_{all}$,
- $C$ is the base image size, which depends on the data set,
- $c$ is the CNN ingestion size, such that $C > c$,
- $L$ is the number of zoom levels that will be applied.

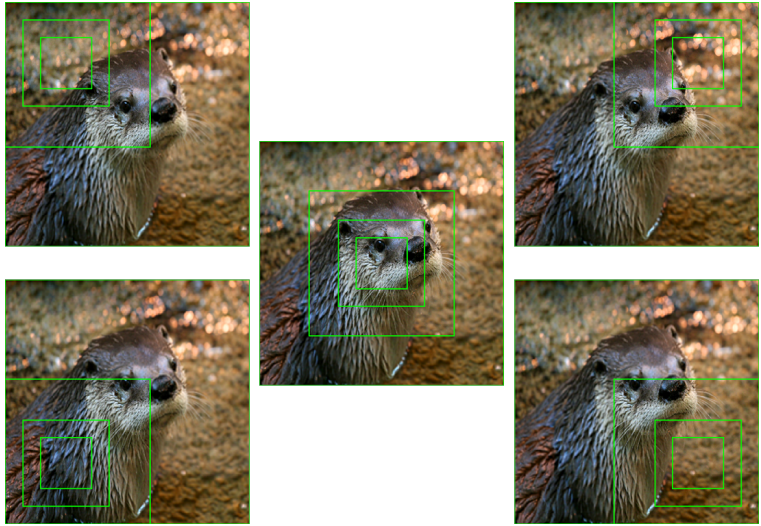

Figure 2: An image of an otter, courtesy of OpenImages [4]. All five of the focal points are depicted with a 4-level zoom implementation. During focal point training, each green box represents an independent image of size 224x224 (including the full image), resulting in 16 unique 224x224 images. During focal point unification, all of the green box estimates for a given focal point are unified to produce a single estimate. In the final step, focal point merging, a collection of unified focal point estimates are merged into a single estimate.

### 3.1 Baseline

The baseline CNN can be any existing CNN with any output layer activation function (sigmoid, softmax, etc). The model can be pre-trained or new, but it is useful to have a pre-trained model to compare pre-ONN accuracy with post-ONN accuracy. The CNN can also have any number of techniques applied to it, such as data augmentation and regularization; the ONN architecture acts as an additional layer on top of the others.

We have selected the VGG-16 model [7] as our baseline due to its proven track record. A "stock" version of the model is used as a baseline, with the only modification being a dropout of 0.5 after each max pooling layer. A stochastic gradient descent (SGD) optimizer is used with a learning rate of 0.001.

The baseline test only runs against images of size $c$, capturing the CNN's potential against images as large as the ingestion size (which is how most CNNs operate).

### 3.2 Focal Point Training

The baseline CNN (pre-trained or new) is then trained against the focal points of the images. Our implementation makes use of 5 known focal points, one of which is selected at random during each training epoch when the image is selected to be trained against. During validation, all focal points are utilized.

While training, a random focal point $\gamma \in F$ is selected for each image. The image is then broken out into multiple zoom levels, forming an image pyramid centered on the focal point. Our ONN implementation uses an exponential zoom coefficient to determine the size of each level:

$$r = round(c \times z^{l-1})$$

where $r$ is the resolution of the downsampled level, $z$ is the zoom coefficient, and $l$ is the current level number (where $l = 1$ is the lowest zoom level and $l = L$ is the highest zoom level). This equation enforces the condition that $r = c$ when $l = 1$, the lowest zoom level. The value of $z$ should be calibrated such that $r = C$ when $l = L$, the highest zoom level.

The base image is then downsampled to each of the levels, using $r$ as the target resolution. From each level, a $c$ sized region centered on $\gamma \times r$ is extracted. If the generated extraction region is not contained within the image boundary, the center is shifted in that level to keep the extraction fully within the resized image. As a result, the lowest zoom level, which is size $r = c$, will be extracted in full regardless of $\gamma$.

To speed up the process, the levels are downsampled in reverse, starting with $l = L$ where no downsampling is needed (as this is the base image). That image is used to generate $l = L - 1$, which is then used down generate $l = L - 2$, etc. If using a static $z$, $c$, and $C$, which is recommended, then the baseline image can be downsampled and cached in preprocessing. We employed an antialias based downsampler to increase generated image quality.

For core CNN training, each of the extracted images are considered independent. This technique therefore yields a $L \times$ increase in the data set size for a given epoch. If all focal points in $F$ are used (either in one epoch or across multiple), then the data is augmented by a factor of $L \times |F|$.

The weights that produced the lowest validation loss are used, but it is worth noting that the validation accuracy at this stage may be lower than the baseline due to all of the zoom regions being included in the measurements. The model is then run against the baseline images again (only the images of size $c$) to allow a direct comparison to the baseline results. The core CNN trained variant should yield a clear accuracy improvement.

### 3.3 Focal Point Unification

With a trained core CNN in hand, the next step in the ONN architecture is to train a network that can classify all zoom levels of a given focal point as a single estimate. The goal of course is that the generated estimate draws from more data, and thus will yield a higher accuracy than just using the baseline images of size $c$.

The implementation style we chose for this segment is a Siamese core CNN arrangement, where each zoom level is pushed through the core CNN in parallel. We then take the results of the final hidden layer (dropping the softmax output layer) and combine the outputs into a single array of size $4096 \times L$. The array is then flattened into a single vector, which is pushed into a new unification neural network, which is responsible for making a single classification estimation.

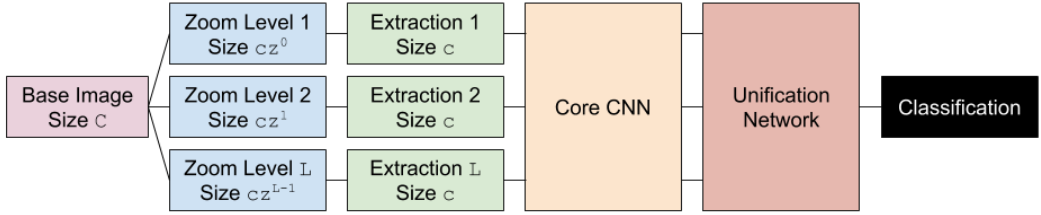

Figure 3: The focal point unifier is broken out into 5 steps. First the base image of size $C$ is broken out into rescaled images of size $round(cz^{l-1})$. Next a $c$ sized region is extracted from each image centered on the provided focal point. Each of the images is run through the CNN in a Siamese fashion, and the values of the last hidden layer for each image are unified by way of the unification network, producing a single classification estimation. During focal point merging, this algorithm is run for multiple focal points, and the classification estimations are combined to generate a single estimate.

Our implementation is a simple two layer neural network with one hidden layer of size 2048 employing ReLU activations, and an output layer that is a carbon copy of the baseline VGG-16 output layer (the number of neurons is equal to the number of classes, and in our case a softmax activation function is used). We employed a high dropout of 0.75 before and after the hidden layer.

Since the core ONN is no longer being trained at this point, we can cache the results for each image and focal point pairing. The unifier network can then be trained relatively quickly against all focal points. The model with the lowest validation loss is kept as the final model, and the accuracy at this point should yield an improvement over the core CNN accuracy.

### 3.4 Focal Point Merging

When evaluating the focal point unification phase, all of the focal points are run against the test images, where each focal point is considered independent. This yields the focal point's unified accuracy. The results of multiple focal points can then be merged into a single estimate, and should provide a small bump over the use of a single focal point.

Our implementation takes the average of all focal point output vectors for a given image, seeking classes that have the strongest support across all of the focal points. This technique can help fight against high-confidence false positives that might occur on a given focal point, but not on the other focal points.

We take the average of all of the focal points in $F$, but a subset can be utilized when using a larger $F$.

## 4 Experiments

For all of the experimentation, no standard data augmentation techniques are utilized. A stronger baseline may have been achievable had we employed image classification techniques such as contrast normalization, minor image resizing and rotation, image flipping, and noise injection, but not all CNN models make use of these techniques. The goal of these experiments is to show that the ONN architecture acts as a powerful generalization tool on its own accord, though advanced techniques can be still be layered on top of the architecture. As a result, we elected not to employ those techniques in the experimentation to maintain clarity in the ONN architecture's results.

Since the VGG-16 model [7] employs a standard input size of (224, 224), we set $c = (224, 224)$. The goal was to perform experiments using $L = 8$, so we selected a simple $z_8 = 1.25$ which yielded $C = (1068, 1068)$. To reuse the same $C$ in the $L = 4$ experiments, we use $z_4 = 1.683$. This yielded sizes $[224, 377, 634, 1068]$ for $L = 4$, and sizes $[224, 280, 350, 438, 547, 684, 854, 1068]$ for $L = 8$.

For all experiments, we use $F = [(0.5, 0.5), (0.25, 0.25), (0.25, 0.75), (0.75, 0.25), (0.75, 0.75)]$.

### 4.1 OpenImages Dataset

Many image sets are of size $C = (224, 224)$ or smaller, but we desired images of size $C = (1068, 1068)$. To meet these demands, we made use of the OpenImages dataset [4]. The OpenImages database contains over 9 million images, with about 5,000 trainable classes.

Two test sets of data were downloaded: one with 100 random classes, and one with 1,000 random classes. Since most of the images bore multiple class labels, one of the available classes was selected randomly for each image. A limit was then placed on the images such that one class could not contain any more than 1,000 images, a necessity as a small handful of classes contained tens of thousands of images, which would artificially inflate the accuracy of the results (especially the top-5 results).

The images were downloaded and resized to 1068x1068, allowing for upscaling of smaller images, and forcing the dimensions even if the images were of a different aspect ratio. Downloaded images were scrubbed for missing images, "Image Not Found" placeholder images, and corrupt files. In the case of the 100 class set, about 8% of the images were scrubbed, yielding 63,127 trainable images. For the 1,000 class set, about 3% of the images were scrubbed, yielding 248,838 images.

The difficulty of the OpenImages dataset makes inclusion of the top-5 test accuracy critical, as the regular accuracy values are fairly low. We felt the need to include both values, as it produces a better overall understanding of model performance.

### 4.2 Baseline

The baseline VGG-16 CNN is trained against 224x224x3 rescaled versions of the 1068x1068x3 files. An SGD optimizer is used with a learning rate of 0.001, and the model was run using early stopping when a clear validation divergence was noted. The 100-label and 1,000-label sets each had their own baseline model. Baseline training thus yielded results of 28.90% accuracy and 65.56% top-5 accuracy for the 100-label set, and 8.65% accuracy and 25.55% top-5 accuracy for the 1,000-label set.

### 4.3 Focal Point Training

For both 4 zoom level and 8 zoom level variants, focal point training uses the baseline weights as a starting point. This provides a reduced training time over random initializations, but also emulates how ONN's can be applied: as a tool to improve generalization of pre-trained models. Experiments with training from random initializers yielded roughly similar results as using pre-trained models, but took longer to converge.

The same SGD optimizer and learning rate are used during this portion of the training. In any given epoch, all of the images are iterated, and a random focal point is selected from the available focal points. That focal point is expanded into all of the component extraction zones, and the resultant extractions are treated as independent images. While it is possible to force all of the focal points into one epoch, the resultant epochs become very large, and thus the validation loss is measured very infrequently. Too infrequent of a measurement for validation loss means less granularity when selecting the best model through early stopping.

Focal point training resulted in a significant improvement in the generalization of the model, yielding a 10-22% reduction in error rates and 27-59% reduction in top-5 error rates.

### 4.4 Focal Point Unification

Unification is a two-step process. First, the models from focal point training are run against every image and focal point. In each case, the model is run against all generated zoom levels, and the resultant 4096 arrays (the results of the final hidden layer are captured as opposed to the output values) are combined and flattened into a single vector. Results are cached for rapid retrieval to expedite training.

Upon completion, the unification network is trained against all of the cached CNN results. Training this network is much faster, and tended to converge in only a few epochs before validation loss divergence. Focal point unification yielded a 14-29% reduction in error rates and 36-68% reduction in top-5 error rates.

| | Optical Neural Network Results | | | | | | | |
|---|---|---|---|---|---|---|---|---|
| | 100 Labels | | | | 1,000 Labels | | | |
| | 4 Zoom Levels | | 8 Zoom Levels | | 4 Zoom Levels | | 8 Zoom Levels | |
| | Accuracy | Top-5 | Accuracy | Top 5 | Accuracy | Top 5 | Accuracy | Top 5 |
| Baseline | 28.90% | 65.56% | 28.90% | 65.56% | 8.65% | 25.55% | 8.65% | 25.55% |
| Focal Point Training | 43.00% | 83.40% | 44.85% | 85.85% | 18.08% | 46.29% | 18.00% | 45.47% |
| Focal Point Unification | 47.40% | 87.89% | 49.51% | 89.04% | 21.89% | 52.72% | 22.16% | 53.47% |
| Focal Point Merging | **49.85%** | **89.79%** | **52.72%** | **91.17%** | **23.53%** | **54.99%** | **24.14%** | **56.18%** |

Table 1: In all experiments, each additional ONN module increases the accuracy and top-5 accuracy by a considerable amount over the baseline. Going from $L = 4$ to $L = 8$ tended to improve the model results, but only to a relatively small degree.

### 4.5 Focal Point Merging

The unification network was run against all of the test data again, but all focal points for a given image were clustered into groups. The outputs from each group were averaged together to produce a single estimation. The accuracy and top-5 accuracy were then measured in the same fashion as the prior experiments, but against the grouped estimations instead of the individual estimations. Merging in this way further improved the accuracy, yielding a 16-34% reduction in error rates and 40-74% reduction in top-5 error rates.

Ultimately the 100 label set netted a 77% increase in accuracy, while the 1,000 label set attained a 175% accuracy gain, both significant improvements.

## 5 Alternative Implementations and Notes

The ONN model demonstrated thus-far has been largely a proof of concept. It shows that making use of ONN architecture has potential for huge improvements in image classification accuracy.

The focal point merging strategy requires a handful of focal points to be run on the image. Our implementation only uses 5 focal points, but larger focal point allowances would need to intelligently select which focal points to leverage instead of using all of them. An example of such a strategy may start with, say, 5 focal points (perhaps the same ones we've used thusfar). The results would be analyzed by a mapping algorithm which would try to predict what the best follow-on focal points might be. This process would then repeat some number of times or until a sufficiently high level of confidence is reached. Alternatively this process could be done in reverse, having an algorithm generate regions of interest on the image, from which the focal points with the highest likelihood of a good prediction will be selected.

The focal point unifier could very likely be improved dramatically. Instead of harvesting data from the final hidden layer of the core CNN model, data could be harvested directly from the final convolution, from the output, or otherwise. Further, how the data is combined might not be optimal, as a dense neural network locks in the number of zoom levels. A better option might be a neural network that operates only on two layers. Starting from the highest zoom level, each level would be "folded into" the level above it. This process would repeat until only one one level remains, which would then be pushed into a post-processor to generate a single classification estimation. This solution would induce some runtime overhead, but would allow a dynamic number of zoom levels to be used during runtime and allow learning has-a relationships between pairs of levels. The folding technique could even allow learning very deep has-a relationships, and thus make very advanced object classification decisions.

A major strength of ONNs is that a degree of scalar invariance is introduced. Thus objects can be classified well even if they only account for a small portion of the provided image, or if they're made up of sub-features that often appear in multiple or varying sizes. The degree of scalar invariance also allows ONNs to be paired well with sliding window approaches for bounding box generation, allowing very poor bounding boxes to still provide high confidence object classification estimations.

The ONN approach in theory produces fairly strong protection against many adversarial techniques. Standard adversarial techniques attempt to inject noise to negatively influence the model and thus produce high confidence yet incorrect estimations. A full ONN uses multiple zoom levels against the same CNN, so generating noise that produces the correct results across all of the levels could be much more difficult. The problem is amplified as multiple focal points are introduced to solve the problem, as with the ONN focal point averaging technique. Though this theory is currently untested, there is a real possibility that ONNs will have high resilience to adversarial techniques, making them an incredibly attractive candidate for implementation in safety centric projects (such as street sign identification for autonomous vehicles).

By removing the softmax at the end of the algorithm, it is easy to envision an algorithm that is trained on has-a instead of is-a. That is, an image would return a 1 for a bottle if it is thought to contain, anywhere in the image, a bottle. Without the softmax on the output layer, any or all of the classes could be contained within any given image. The ONN training would be highly suited for this approach, as the first level image could be processed to generate a list of assumed objects within the image. From there, another algorithm could determine "hot spots" within the image based on the estimated has-a classes. Focal points would be generated and the process would be repeated for zoomed in sub-regions, forming a hierarchical object search. The same CNN would be used at every zoom level, allowing large batches of focal points to be run at once to make the algorithm highly parallelized. In theory, this style of algorithm could be applied to images of any input size, using one CNN to identify objects of any size throughout the image.

An alternative architecture that avoids multiple CNN runs and a separated unifier network (at the cost of having to lock in the number of layers a priori) is to use a single data volume for the extracted images for a given focal point. This would require a different approach to the underlying CNN, as a VGG-16 implementation with 4 layers would require a 224x224x12 input volume, changing the dynamics of the processor. This technique, however, could allow extremely robust cross-zoom-level feature identification and provide a more streamlined model with fewer moving pieces.

# 6 Conclusion

By emulating the loss in resolution of the human eye as distance from the focal point increases, optical neural networks are able to make use of multiple zoom levels to increase object classification accuracy. Features can be learned at any number of scales, allowing them to be readily identified in a wider range of configurations and environments.

Focal points on an image can be used to produce multiple levels of zoom, which can be used as a form of data augmentation for the underlying CNN. In all experiments, focal point training yielded a significant reduction in error and top-5 error rates over the baseline model.

A unification model can then be constructed to combine each zoom level of a focal point into a single estimation, yielded further improvements over focal point training. As a final element, multiple focal points can be merged into a single estimation, again reducing error rates. While utilizing all elements of the architecture, significant reductions in error rates can be achieved, and even more significant reductions in top-5 error rates.

The ONN architecture as shown in this paper is just a proof of concept implementation, yet has achieved noteworthy error reductions. The modular design opens up the architecture for significant algorithmic improvements through further research.

There are many potential applications for the ONN framework, which can be implemented modularly as needed for the problem at hand. Even if just used as a data augmentation technique for an existing network, the ONN framework has the potential to provide notable accuracy improvements.